\documentclass{bmvc2k}
\usepackage{multirow}
\usepackage{array}
\usepackage{amssymb}
\usepackage{booktabs}
\newcommand\blfootnote[1]{%
  \begingroup
  \renewcommand\thefootnote{}\footnote{#1}%
  \addtocounter{footnote}{-1}%
  \endgroup
}

\title{Talking Head Generation with Audio and Speech Related Facial Action Units}

\addauthor{Sen Chen}{chensen@tju.edu.cn}{1}
\addauthor{Zhilei Liu$^\star$}{zhileiliu@tju.edu.cn}{1}
\addauthor{Jiaxing Liu}{jiaxingliu@tju.edu.cn}{1}
\addauthor{Zhengxiang Yan}{zhengxiangyan@tju.edu.cn}{1}
\addauthor{Longbiao Wang$^\star$}{longbiao_wang@tju.edu.cn}{1}

\addinstitution{
 College of Intelligence and Computing \\
 Tianjin University\\
 Tianjin, 300072, China\\
}

\runninghead{Chen et al.}{Talking Head Generation with Audio and Speech Related AUs}

\begin{document}
\maketitle
\blfootnote{\hspace{-0.55cm} *Corresponding Authors.}

\begin{abstract}
The task of talking head generation is to synthesize a lip synchronized talking head video by inputting an arbitrary face image and audio clips. Most existing methods ignore the local driving information of the mouth muscles. In this paper, we propose a novel recurrent generative network that uses both audio and speech-related facial action units (AUs) as the driving information. AU information related to the mouth can guide the movement of the mouth more accurately. Since speech is highly correlated with speech-related AUs, we propose an Audio-to-AU module in our system to predict the speech-related AU information from speech.   In addition, we use AU classifier to ensure that the generated images contain correct AU information. Frame discriminator is also constructed for adversarial training to improve the realism of the generated face. We verify the effectiveness of our model on the GRID dataset and TCD-TIMIT dataset. We also conduct an ablation study to verify the contribution of each component in our model. Quantitative and qualitative experiments demonstrate that our method outperforms existing methods in both image quality and lip-sync accuracy.
\end{abstract}
\section{Introduction}

Recently, talking head generation has attracted more and more attention in the fields of academic and industry, which is essential in the applications of human-computer interaction, film making, virtual reality, computer games, etc. This research explores how to generate a talking head video by inputting anyone's image as an identity image and driving information related to mouth movement, e.g., speech audio, and text. 

Before deep learning became popular, many researchers in early work relied on Hidden Markov Models (HMM) to capture the dynamic relationship between audio and lip motion \cite{bregler1997video,yamamoto1998lip,xie2007coupled}. In recent years, Deep Neural Networks (DNN) is widely used in talking head generation. Some methods selected the best-matched lip region image from the specific person's database by inputting audio information, then synthesize it into the target face \cite{journals/tog/SuwajanakornSK17,fan2015photo}. These methods are subject dependent and bring a lot of overhead when transferring to a new subject. Later, many works study how to generate arbitrary speaker instead of specific speaker \cite{DBLP:conf/aaai/Zhou000W19,chung2017you,prajwal2020lip}. However, these works ignore the temporal relationship of features, so the generated videos are accompanied by jitter. Some researchers used sequence model to learn the temporal relationship to reduce jitter \cite{song2018talking,vougioukas2019realistic,eskimez2020end}. Nevertheless, talking is also a kind of movement driven by facial muscles, especially in the mouth region. Existing methods ignore the local driving information of the mouth muscles, resulting in the lip movement are not very synchronized with the audio. Inspired by this limitation, we consider using both audio and local information of mouth muscles to drive the talking head generation. 

Facial Action Coding System (FACS) is a comprehensive and objective description system of facial movements~\cite{Ekman}. It defines a set of basic facial action units (AUs), and each AU represents a basic facial muscle movement. The value of AUs can either use binary classification to indicate whether these AUs are activated, or use intensity value to indicate activation intensity. FACS has attracted much attention in face editing \cite{pumarola2018ganimation,li2019face,liu2020region}, such as facial expression editing \cite{pumarola2018ganimation}. These works proved that AU information can be used to edit local facial regions. AU labels are usually extracted by face image detection \cite{baltrusaitis2018openface,shao2018deep}, but they can also be extracted through other modalities related to facial motion, such as speech audio. A few researchers have constructed the relationship between audio information and speech-related AU information in recent years \cite{meng2017listen,ringeval2015face}. They proved that audio as information for speech-related AU recognition is feasible. Therefore, we propose an Audio-to-AU module to obtain speech-related AU representation from speech as local driving information.

To tackle the limitations of existing methods that ignore the local information of the mouth muscles, this paper proposes a novel talking head generation model using speech-related AUs as local information to drive the muscle movements in the mouth region. Specifically, we use both audio and AU representation as the driving information, so that audio information drives the whole mouth, and AU information focuses on the local muscles. In addition, we also use a recurrent generative network to capture the temporal dependence between consecutive frames. In summary, our contribution can be summarized as follows:
\begin{itemize}
\item A novel recurrent generative network is proposed by integrating both audio and speech-related facial AUs as driving information to drive talking head generation. 

\item An Audio2AU module is designed in our framework to predict speech-related AU information from speech. Moreover, we add pre-trained AU classifier to supervise that the generated images contain correct AU information.

\item Extensive quantitative and qualitative experiments conducted on two datasets demonstrate that our framework achieves high-quality talking head generation of arbitrary identities and achieves significant improvement over existing methods.
\end{itemize}
\section{Related Work}
\subsection{Audio-driven talking head generation}
In early work, researchers relied on the modeling method of HMM. Bregler et al.\cite{bregler1997video} proposed a video rewrite method to match the lip movement to the phoneme sequence of the new audio track. Xie et al.\cite{xie2007coupled} used a coupled Hidden Markov Model (cHMM) to model the subtle relationship of audio-visual speech. Later, deep learning methods were introduced to talking head generation. Suwajanakorn et al. \cite{journals/tog/SuwajanakornSK17} used a database of speeches by the then US President Obama to retrieve the lip region image that best matches the input audio, and synthesized it into the target image. Chung et al. \cite{chung2017you} proposed an encoder-decoder structure that uses audio frames to drive the mouth movement of the input face. Prajwal et al. \cite{prajwal2020lip} used a pre-trained SyncNet as lip-sync discriminator for adversarial training with the generator to supervise the correctness of mouth motion. Eskimez et al. \cite{eskimez2021speech} introduced an emotion encoder to generate talking face video with a specific emotion, while our method generates talking face video with specific AUs. Zhou et al. \cite{zhou2020makelttalk} disentangled the content and speaker information in the input audio signal, and used landmarks as an intermediate representation that reflect speaker-aware dynamics. Different from the method of Zhou et al. \cite{zhou2020makelttalk}, we use speech-related AUs to indicate whether the specific muscles should be activated, instead of learning the location information represented by landmarks. Some researchers use the sequence modeling method to enhance the temporal dependence of generated frames. Song et al. \cite{song2018talking} combined audio features and identity features into a Recurrent Neural Network (RNN) to enhance the temporal dependence. Vougioukas et al. \cite{DBLP:conf/bmvc/VougioukasPP18} proposed a temporal Generative Adversarial Network (GAN) to generate more natural face image sequence. However, all the above methods did not consider the local information of the mouth muscles.

\subsection{AU-based face editing}
As one of the most comprehensive ways to describe facial movements, Facial Action Coding System (FACS) has recently attracted widespread attention \cite{martinez2017automatic,pumarola2018ganimation,liu2020region}. Pumarola et al. \cite{pumarola2018ganimation} proposed an AU-based face editing system, which uses AU intensity labels to edit the input face to generate a face with specific facial muscles action. Liu et al. \cite{liu2020region} utilized local AU regional rules to control the status of each AU and used an attentive mechanism to combine them into the whole facial expressions. Zhou et al. \cite{zhou2017photorealistic} proposed a conditional difference adversarial autoencoder for facial expression synthesis, which can generate a face image with a target AU label. In our method, we utilize speech-related AU information to edit the local region of the mouth.
\subsection{Speech-related AU recognition}
Since speech is highly correlated with speech-related AUs, a few researchers have tried to establish the relationship between audio information and speech-related AU information in recent years \cite{meng2017listen,ringeval2015face}. Meng et al. \cite{meng2017listen} proposed a continuous-time Bayesian network (CTBN) to simulate the dynamic relationship between phoneme and AUs. Then AU recognition is performed by probabilistic inference via the CTBN model. Ringeval et al. \cite{ringeval2015face} used LLD features of acoustic feature datasets to carry out action unit recognition through an RNN-LSTM model. In our work, we propose an Audio-to-AU module to predict speech-related AU information from every audio frame to guide more accurate mouth movement.
\subsection{GAN-based image synthesis}
Generative Adversarial Network (GAN) \cite{goodfellow2014generative} is a framework for estimating generative models through adversarial processes. It includes a generator to produce realistic fake samples and a discriminator to distinguish real and fake images. Recently works have proved that GAN can produce realistic images, and it has achieved excellent results in many fields, such as image translation \cite{isola2017image}, face generation \cite{chen2019hierarchical,zhou2021pose,song2018talking}. Chen et al. \cite{chen2019hierarchical} devised a cascade GAN approach to generate the talking face. Eskimez et al. \cite{eskimez2020end} used GAN training to improve the image quality and mouth-speech synchronization. Here, we use the GAN training method to enforce the generated image distribution to approach real image distribution.
\section{Proposed Method}

\begin{figure*}[t]
\vspace{-1.5mm}
  \centering
  \includegraphics[width=\linewidth]{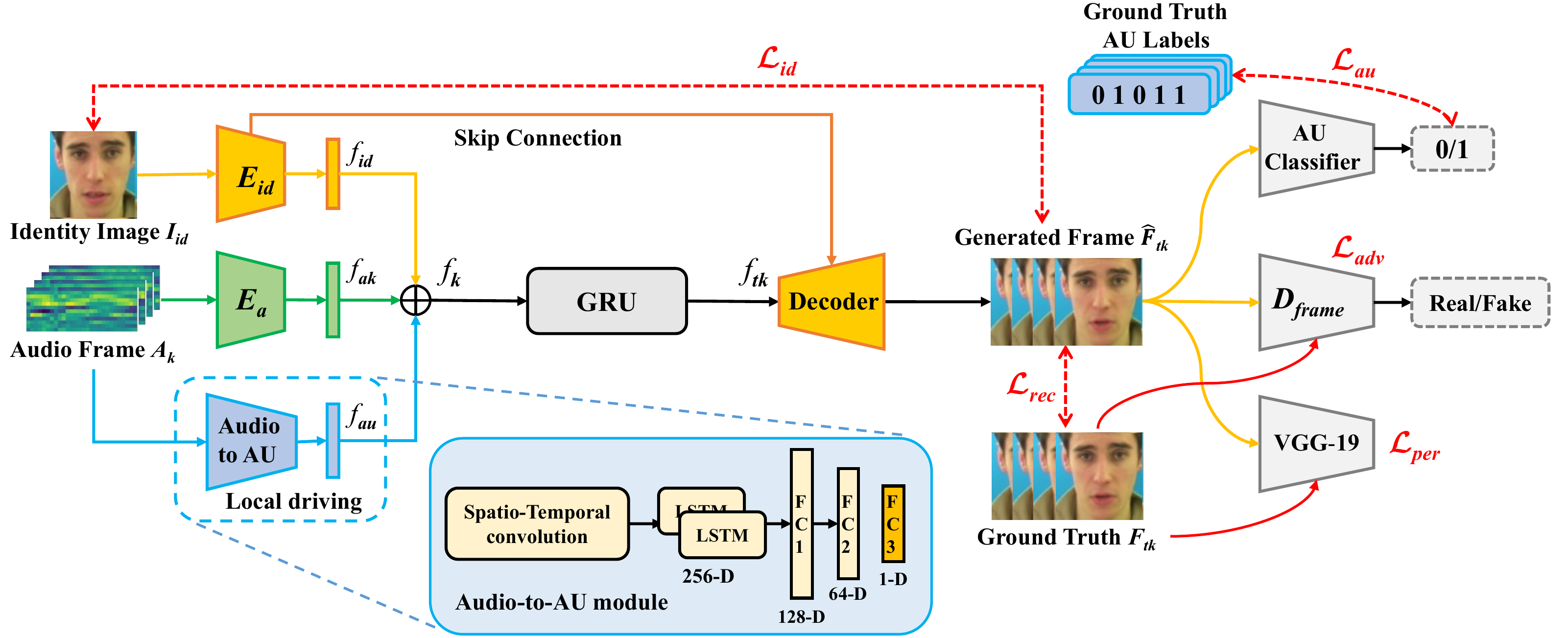}
  \caption{The pipeline of our proposed method}
  \vspace{-2.5mm}
  \label{fig.1}
\end{figure*}

The architecture of our proposed method is shown in Fig.~\ref{fig.1}. The generator contains an identity encoder, audio encoder, Audio-to-AU module, RNN module, and image decoder. In addition, we propose AU classifier to supervise whether the AU information of the generated image is accurate. We use the VGG-19 network \cite{simonyan2014very} to extract high-level features of the generated frame and ground truth frame, and compare the differences between them. Frame discriminator is used for adversarial training \cite{goodfellow2014generative} with the generator to make the generated frame more realistic. In the following, we describe each module in detail.

\subsection{Network architecture}

\textbf{Identity Encoder:} The identity encoder \(E_{id}\) contains four 2D convolution layers and a fully connected layer, each convolution layer is followed by ReLU activation function. The input \(I_{id}\) is a face image resized to 112 \(\times\) 112 as identity image. It can be a randomly selected frame from the video. For convenience, we use the first frame of each video in this paper. The output of \(E_{id}\) is a 512-dimensional identity feature vector \(f_{id}=E_{id}(I_{id})\).

\noindent\textbf{Audio Encoder:} The audio encoder \(E_{a}\) contains five 2D convolution layers and two fully connected layers, and each convolution layer is followed by batch normalization and ReLU activation function. For an audio sequence, we preprocess it before input. Specifically, we extract Mel-Frequency Cepstral Coefficient (MFCC) features, and then use a fixed size sliding window to crop MFCC segments continuously. Finally, the input are continuous MFCC frames \(A=[A_{1}, A_{2}, ..., A_{k}, ..., A_{n}]\), and the outputs are a series of 512-dimensional audio feature vectors \(f_a=[f_{a1}, f_{a2}, ..., f_{ak}, ...,  f_{an}]\), where \(f_{ak}=E_{a}(A_{k})\).

\begin{figure}[t]
  \centering
  \includegraphics[width=0.6\linewidth]{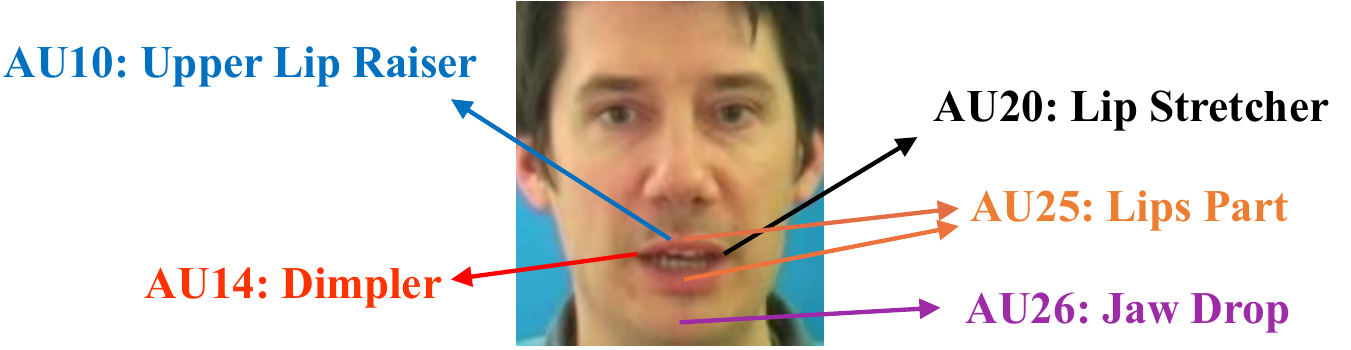}
  \caption{Interpretation of speech-related AUs}
  \vspace{-2.5mm}
  \label{fig.2}
\end{figure}

\noindent\textbf{Audio-to-AU module:} To guide the muscle movements in the mouth region more accurately, we propose an Audio-to-AU module to extract speech-related AU information from the speech in real-time. It is pre-trained with the paired audio and AU data. The pre-trained Audio-to-AU module is shown in Fig.~\ref{fig.1}. In the pre-training stage, the input MFCC features are respectively convolved in the frequency domain and time domain. They then pass through two LSTM layers and three fully connected layers and use the sigmoid activation function to obtain the probability of AU occurrence. After pre-training, we remove the last fully connected layer of this network as the Audio-to-AU module and add it to our framework, because the multi-dimensional AU representation obtained by the penultimate full connected layer is more conducive to the model to learn AU information than a one-dimensional label. Since talking is a facial movement driven by multiple facial muscles in the mouth region, multiple Audio-to-AU modules are necessary to extract this local facial information. Based on the anatomy knowledge \cite{martinez2017automatic}, five speech-related AUs are selected in this work, namely AU10, AU14, AU20, AU25, and AU26. The detailed interpretation is shown in Fig.~\ref{fig.2}. Each output of Audio-to-AU module is a 64-dimensional AU feature vector \(f_{au_i}\), where \(i\) is the number of the \(i\)-th AU. We concatenate the five 64-dimensional vectors and finally get the 320-dimensional AU feature vector \(f_{au}=f_{au_{10}}\oplus f_{au_{14}}\oplus f_{au_{20}}\oplus f_{au_{25}}\oplus f_{au_{26}}\) .

\noindent\textbf{Recurrent Neural Network module:} The RNN module uses a Gate Recurrent Unit (GRU) layer to maintain the temporal dependence between frames. We concatenate the identity feature \(f_{id}\), the audio feature \(f_{ak}\) and the speech-related AUs feature \(f_{au}\) to get \(f_{k}=f_{id}\oplus f_{ak}\oplus f_{au}\), where $k$ represents the \(k\)-th frame in the sequence. We input the feature sequence into GRU and get \(f_{t}=[f_{t1}, f_{t2}, ..., f_{tk}, ..., f_{tn}]\),
where \(f_{tk} = RNN(f_k)\).

\noindent\textbf{Image Decoder:} The image decoder is used to generate a talking head video. It consists of a fully connected layer and six transposed convolution layers. In order to preserve the input identity information and the facial texture, we utilize a structure similar to U-Net \cite{ronneberger2015u}, which uses skip connection between the identity encoder and the image decoder. From the input feature \(f_{t}=[f_{t1}, f_{t2}, ..., f_{tk}, ..., f_{tn}]\) , we can get the decoded image sequence \( \hat{F_{t}}=[\hat{F}_{t1}, \hat{F}_{t2}, ..., \hat{F}_{tk}, ..., \hat{F}_{tn}]\).

\noindent\textbf{AU Classifier:} The AU classifier is used to predict the occurrence probability of the speech-related AUs in the generated frames. It consists of four convolution layers and three fully connected layers. Every two convolution layers are followed by a max-pooling layer. Then the sigmoid function is used to get the occurrence probability of each AU. To focus on the mouth region, we only input the lower face of the generated image. Binary Cross-Entropy Loss is used to calculate the loss between the predicted AU and the ground truth AU label:
\begin{equation}
  \ \mathcal{L}_{bce} = - \frac{1}{n_{au}}\sum_{i=1}^{n_{au}} w_i[y_i\log \hat{y}_i+(1-y_i )\log(1-\hat{y}_i)]
  \label{eq.1}
\end{equation}
where \(y_i\) represents the ground truth label of the \(i\)-th AU, which is 1 if occurrence and 0 otherwise, and \(\hat{y}_i\) represents the corresponding predicted probability of occurrence. For most facial AU detection benchmarks, the occurrence rates of AUs are imbalanced \cite{martinez2017automatic}. Because AUs are not independent of each other, the imbalance of training data will negatively impact the performance. Therefore, we add weight \(w_i\) in Eq.\ref{eq.1} to alleviate the data imbalance problem, where \(w_i=\frac{(1/r_i)n_{au}}{\sum_{i=1}^{n_{au}}{1/r_i}}\), \(r_i\) is the occurrence rate of the \(i\)-th AU in the training set \cite{shao2018deep}.

Some AUs rarely appear in the training set, so the network prediction tends to be absent. To alleviate this problem, we introduce a weighted multi-label Dice coefficient loss \cite{milletari2016v}: 
\begin{equation}
  \ \mathcal{L}_{dice} = \frac{1}{n_{au}}\sum_{i=1}^{n_{au}} w_i[1-\frac{2y_i\hat{y}_i+\epsilon}{y_i^2+\hat{y}_i^2+\epsilon}]
  \label{eq2}
\end{equation}
where \(\varepsilon\) is the smooth term. Thus, our AU loss is defined as:

\begin{equation}
  \ \mathcal{L}_{au} = \mathcal{L}_{bce} +\mathcal{L}_{dice} 
  \label{eq3}
\end{equation}

\noindent\textbf{Frame Discriminator:} The frame discriminator is used for adversarial training to improve the realism of generated frame \(\hat{F}_{tk}\). The adversarial loss is shown in Eq.\ref{eq.4}:
\begin{equation}
   \mathcal{L}_{adv} =\mathbb{E}_{F_{tk}\sim P_F}[\log D_{frame}(F_{tk})] +  \mathbb{E}_{\hat{F}_{tk}\sim P_{\hat{F}}}[\log (1-D_{frame}(\hat{F}_{tk}))]
  \label{eq.4}
\end{equation}
where \(P_F\) and \(P_{\hat F}\) denote the distributions of ground truth images and generated images respectively, and \(F_{tk}\) is the ground truth of the \(k\)-th frame.

\noindent\textbf{VGG-19 Network:} In order to improve the quality of generated images, we use perceptual loss \cite{johnson2016perceptual} to reflect perceptual-level similarity of images. The pre-trained VGG-19 network \cite{simonyan2014very} is adopted as the perceptual feature extractor. The perceptual loss is defined as:
\begin{equation}
  \ \mathcal{L}_{per}(F_{tk},\hat{F}_{tk}) = \frac{1}{n}\sum_{i=1}^n ||\phi_i(F_{tk}) - \phi_i(\hat F_{tk}) ||_1
  \label{eq5}
\end{equation}
where \(\phi_i\) denotes the \(i\)-th feature extraction layer of the VGG-19 network. 

\subsection{Loss functions}

In addition to AU loss, adversarial loss and perceptual loss mentioned above, reconstruction loss and identity loss are also considered. The reconstruction loss \({L}_{rec}\) is used to minimize the pixel-level difference between the generated image \(\hat{F}_{tk}\) and the real image \(F_{tk}\):
\begin{equation}
  \ \mathcal{L}_{rec}(\hat{F}_{tk}, F_{tk}) =  ||\hat F_{tk} - F_{tk} ||_1
  \label{eq6}
\end{equation}

The identity loss \({L}_{id}\) is used to maintain the identity information, and to reduce the jitter effect of generated video. Specifically, it penalizes the difference between the upper face of \(\hat{F}_{tk}\) and the upper face of \(I_{id}\):

\begin{equation}
  \ \mathcal{L}_{id}(\hat F_{tk_p}, I_{id_p})=\frac{1}{W\times\frac{H}{2}}  \sum_{p\in [0,W]\times[0,\frac{H}{2}]}||\hat F_{tk_p} - I_{id_p} ||_1
  \label{eq7}
\end{equation}
where $W$ and $H$ are the width and height of each frame respectively, \(\hat{F}_{tk_p}\) is the upper half of generated image \(\hat{F}_{tk}\), and \(I_{id_p}\) is the upper half of input image \(I_{id}\).

Finally, the overall loss of our proposed framework is defined as Eq. \ref{eq8}:
\begin{equation}
  \ \mathcal{L}_{total} =\lambda_{rec} \mathcal{L}_{rec} +\lambda_{id} \mathcal{L}_{id} +\lambda_{per}\mathcal{L}_{per} +\lambda_{au}\mathcal{L}_{au} +\lambda_{adv}\mathcal{L}_{adv} 
  \label{eq8}
\end{equation}
where \(\lambda_{rec}\), \(\lambda_{id}\), \(\lambda_{per}\), \(\lambda_{au}\), \(\lambda_{adv}\) are trade-off parameters.
\section{Experimental Results}

\subsection{Dataset}

We conduct extensive experiments on the GRID dataset \cite{cooke2006audio} and TCD-TIMIT dataset \cite{harte2015tcd}. The GRID dataset is a large audio-visual corpus, which consists of high-quality audio and video recordings of 1000 sentences spoken by each of 33 speakers. The TCD-TIMIT dataset has high-quality audio and video footage of 59 speakers uttering approximately 100 phonetically rich sentences each. In our experiments, the GRID dataset is divided into the training set and test set at the ratio of 8: 2, with 27 speakers as the training set and 6 speakers as the test set. For the TCD-TIMIT dataset, we divide 49 speakers into the training set and 10 speakers into the test set. We extract all the video frames and use the Dlib toolkit \cite{king2009dlib} of the HOG-based face detection algorithm to detect all the faces. Then we crop the face regions and resize them to 112\(\times\)112. OpenFace \cite{baltrusaitis2018openface,baltruvsaitis2015cross} is used to detect the AU labels of each real face image as ground truth labels. For the audio inputs, we extract 12-dimensional MFCC features (excluding energy dimension). We try different lengths of audio frames and find that 280ms performs best. Then we align the middle of each audio frame with a corresponding video frame. The audio sliding window slides synchronously with the video frame. 

\subsection{Training details}

Our network is implemented using Pytorch and trained on a single NVIDIA Titan V GPU. The Audio-to-AU module and the AU classifier are pre-trained on the GRID dataset, and when experimenting on the TCD-TIMIT dataset, they are refined to fit the new dataset. The VGG-19 network is pre-trained on the ImageNet dataset \cite{russakovsky2015imagenet}. We adopt Adam optimizer with \(\beta_1=0.5\) and the learning rate of 0.0002 during training. The weights of  \(\lambda_{rec}\), \(\lambda_{id}\), \(\lambda_{per}\), \(\lambda_{au}\), \(\lambda_{adv}\) are 1.5, 1.5, 0.07, 0.02, 0.002 respectively. We use \(\mathcal{L}_{au}\) to fine-tune the Audio-to-AU module and the AU classifier during training, and adopt Adam optimizer with the learning rate of 1e-6 and 1e-7, respectively. We first train our network without frame discriminator for 20 epochs, then add it to fine-tune the network for another 20 epochs. 

\subsection{Quantitative results}

To evaluate the quality of generated images, we adopt the reconstruction metrics Peak SNR (PSNR) and Structural Similarity (SSIM) \cite{wang2004image}. For the lip-sync performance, we verify the recognition accuracy and F1 score of the five selected speech-related AUs in generated frames. Specifically, we use the OpenFace toolkit \cite{baltrusaitis2018openface,baltruvsaitis2015cross} to detect the state of the five selected AUs (activated or not) in each generated frame, then compare them with ground truth labels. 

\begin{table}[!htb]
\begin{center}
\small
\setlength\tabcolsep{1pt}
\begin{tabular}{|l|c|c|c|c|c|c|c|c|}
\hline
\multirow{2}*{Method}&\multicolumn{4}{c|}{GRID}&\multicolumn{4}{c|}{TCD-TIMIT}\\
\cline{2-9}
&PSNR↑&SSIM↑&Avg. F1&Avg. Acc.&PSNR↑&SSIM↑& Avg. F1&Avg. Acc. \\
\hline\hline
CRAN\cite{song2018talking}& 28.041 & 0.694&0.710 &78.71
&24.381 & 0.650&0.562 &81.41 \\
Speech2Vid\cite{jamaludin2019you}& 28.863 & 0.755&0.738 &78.97
&25.132 & 0.685&0.556 &81.91 \\
Baseline ($\mathcal{L}_{rec}$)  & 28.681 &0.746 &0.725&77.77
&25.003 & 0.717&0.545 &80.96\\
Proposed method & \textbf{29.838} &\textbf{0.769}&\textbf{0.751}&\textbf{80.92}
&\textbf{26.201} & \textbf{0.745}&\textbf{0.590} &\textbf{84.92}\\
\hline
\end{tabular}
\end{center}
\caption{Quantitative results on the GRID test set and TCD-TIMIT test set. Avg. F1 and Avg. Acc. are average F1 score and average accuracy (\%) of speech-related AUs respectively.}
\label{tab:1}
\end{table}

Table \ref{tab:1} shows the quantitative results on the GRID test set and TCD-TIMIT test set. To compare with other methods, we implement the methods proposed by Song et al. \cite{song2018talking} and Jamaludin et al. \cite{jamaludin2019you} in the same conditions, and use the same training-testing data split as our proposed method. Our baseline only uses reconstruction loss. We can see that our proposed method has significant improvements in image quality and lip-sync accuracy in both datasets. Compared with Song et al. \cite{song2018talking}, our proposed method improves PSNR by about 1.8 and average F1 score of AUs by 4.1\% on the GRID dataset. Our method also improves PSNR by 1.82 and average accuracy of AUs by about 3.5\% on the TCD-TIMIT dataset. Similarly, our method is significantly higher than Jamaludin et al. \cite{jamaludin2019you} in all metrics.

\begin{figure}[h]
  \centering
  \includegraphics[width=\linewidth]{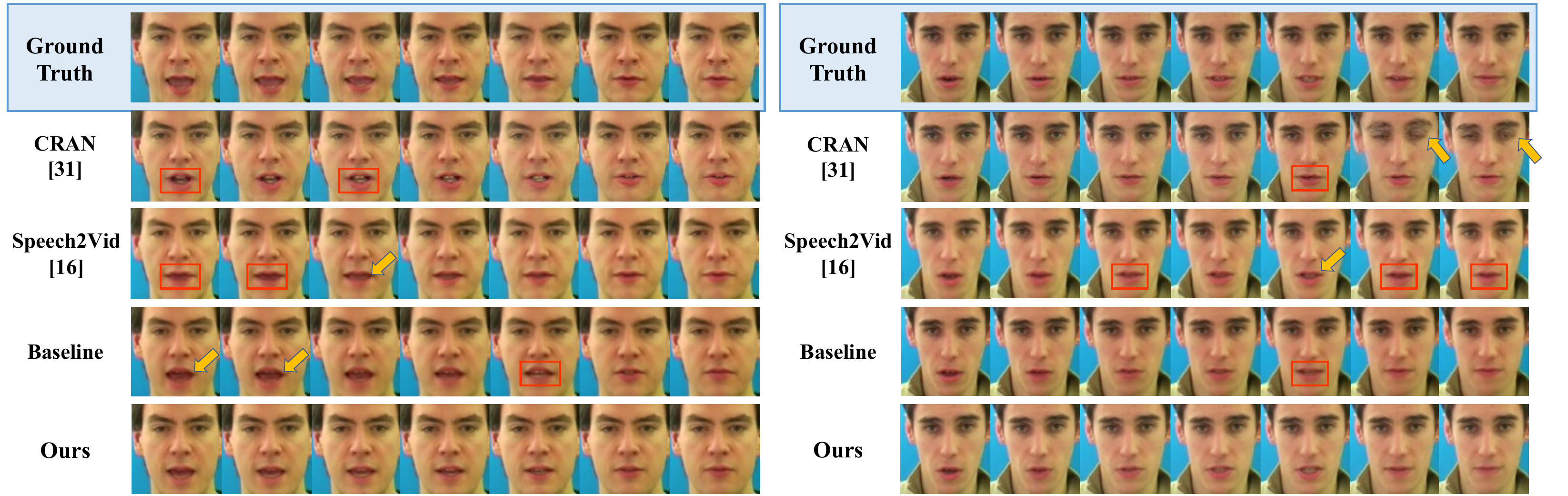}
  \caption{Qualitative results produced by our proposed method and other methods on the GRID test set. Each speaker is a continuous sequence of talking faces. Red boxes are used to mark the examples of lip out of sync, and arrows indicate the examples of blurring.}
  \vspace{-2.5mm}
  \label{fig.3}
  
\end{figure}

\begin{figure}[h]
  \centering
  \includegraphics[width=\linewidth]{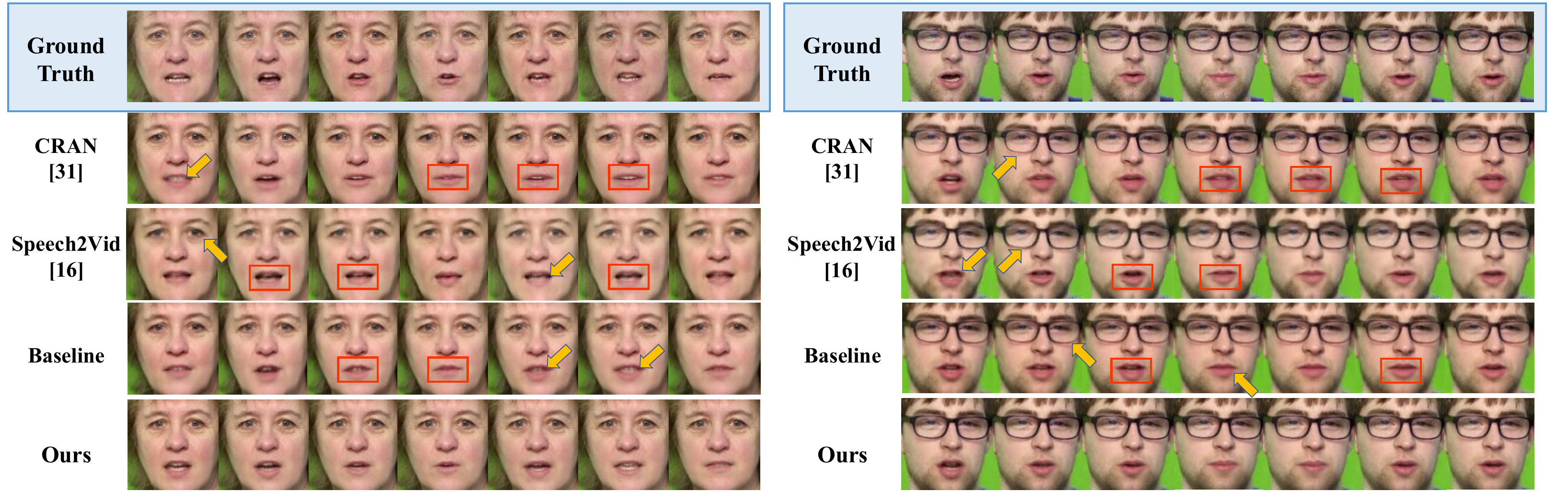}
  \caption{Qualitative results were produced by our proposed method and other methods on the TCD-TIMIT test set.}
  \vspace{-2.5mm}
  \label{fig.4}
\end{figure}

\subsection{Qualitative results}

The qualitative results on the GRID test set are shown in Fig.~\ref{fig.3}. We use two different audio sequences to generate talking head for two speakers. Compared with others, our method performs better in both image quality and lip-sync accuracy. For example, the images generated by Song et al. \cite{song2018talking} has the problem of lip out of sync, such as the mouths marked with red boxes in Fig.~\ref{fig.3}. Besides, there are subtle artifacts in the eye region indicated by the arrows on the right picture, and the generated faces are blurry. Similarly, our baseline and Jamaludin et al. \cite{jamaludin2019you} also has the problem of lip out of sync, and the teeth indicated by the arrows are very blurry. Our proposed method overcomes these shortcomings and performs well. 

To further prove the effectiveness of our model, we also show the qualitative results on the TCD-TIMIT test set in Fig.~\ref{fig.4}. The face images generated by our method are obviously clearer than others. We still use the red boxes to mark the examples of mouth movement out of sync, and use arrows to indicate the examples of blurring, such as teeth. In the left picture, the eyes generated by Song et al. \cite{song2018talking} are a bit dull, and the eyes generated by Jamaludin et al. \cite{jamaludin2019you} are blurry. Besides, the movement of the mouth generated by Jamaludin et al. \cite{jamaludin2019you} is very mismatched. In the right picture, except for the blurred teeth and lips, the arrows also indicate that the glasses generated by other methods are blurry. Our proposed method performs best in both facial texture and mouth movement.

\subsection{Ablation study}

To quantify the contribution of each component of our method, we conduct an ablation study on the GRID dataset. As shown in Table~\ref{tab:2}, all the metrics have been improved after adding perceptual loss, identity loss, adversarial loss, and AU loss progressively. Especially after adding identity loss, PSNR has improved significantly. We think that when AU loss is used together with the Audio-to-AU module, it can achieve the maximum effect. Finally, when we add the Audio-to-AU module, which can promote each other with AU classifier, all the metrics have achieved the best results, especially the average F1 score and average accuracy of AUs have been significantly improved. These experiments prove the effectiveness of our proposed method. To observe the results of each AU, we also show the F1 score and accuracy of the five speech-related AUs respectively in Table \ref{tab:2}. Most of the results of our proposed model perform best. 

\begin{table}[h] 
\begin{center}
 \tiny
  \centering
   \setlength\tabcolsep{4.7pt}
   \begin{tabular}{|l|c|c|c|c| cc|cc|cc|cc|cc |}
  \hline
  \multirow{2}*{Method} &\multirow{2}*{PSNR↑}&\multirow{2}*{SSIM↑}&
    \multirow{1}{*}{Avg.}&\multirow{1}{*}{Avg.}&
                                                                       \multicolumn{2}{c|}{AU10}&
                                                                       \multicolumn{2}{c|}{AU14}&
                                                                       \multicolumn{2}{c|}{AU20}&
                                                                       \multicolumn{2}{c|}{AU25}& 
                                                                       \multicolumn{2}{c|}{AU26}   \\
                                         
 &&&F1&Acc.& F1& Acc. &F1& Acc.& F1& Acc.& F1& Acc. & F1&Acc. \\
  \hline
  \hline
    Baseline ($\mathcal{L}_{rec}$)              & 28.681  & 0.746 &     0.725  & 77.77 &  
    0.745 & 71.14 &  0.477 &82.01 &0.762 &74.86 &0.939 &89.21 &0.703 &71.65\\
    $\mathcal{L}_{rec,per}$           & 28.897  & 0.748   & 0.732  &78.76    &
    \textbf{0.771} & 75.99 &  0.501 &82.95 &0.742 &74.26 &0.940 &89.10 &0.708&71.50\\
   $\mathcal{L}_{rec, per, id}$            & 29.403 & 0.753  &0.735   &79.27   &
   0.757 & 74.87 &  0.516 &82.20 &0.774 &79.13 &0.943 &89.75 &0.685 &70.41\\
    $\mathcal{L}_{rec, per, id, adv}$          & 29.528 & 0.755  &   0.736  &79.55&
   0.756 & 74.40 &  0.511 &83.30 &0.788 &80.37 &0.945 &90.15 &0.682 &69.51\\
    $\mathcal{L}_{rec, per, id, adv, au}$      & 29.618 & 0.764  &   0.740  &80.20&
   0.761 & 75.71 &  0.481 &83.80 &0.800 &80.70 &0.946 &90.17 &\textbf{0.710} &70.63\\
    Full model(+Audio2AU)    & \textbf{29.838}  & \textbf{0.769}  &\textbf{0.751}  & \textbf{80.92} &
   0.763 & \textbf{76.37} &  \textbf{0.531} &\textbf{83.96} &\textbf{0.805} &\textbf{81.64} &\textbf{0.949} &\textbf{90.79} &0.707 &\textbf{71.83}\\
  \hline
  \end{tabular}
  \end{center}
    \caption{Quantitative ablation study on the GRID test set. $\mathcal{L}_{rec}$, $\mathcal{L}_{per}$, $\mathcal{L}_{id}$, $\mathcal{L}_{adv}$, $\mathcal{L}_{au}$ are reconstruction loss, perceptual loss, identity loss, adversarial loss and AU loss, respectively. Avg. F1 and Avg. Acc. are average F1 score and average accuracy (\%) of AUs respectively. }
  \label{tab:2}
\end{table}

\begin{figure}[t]
\centering
  \includegraphics[width=0.62\linewidth]{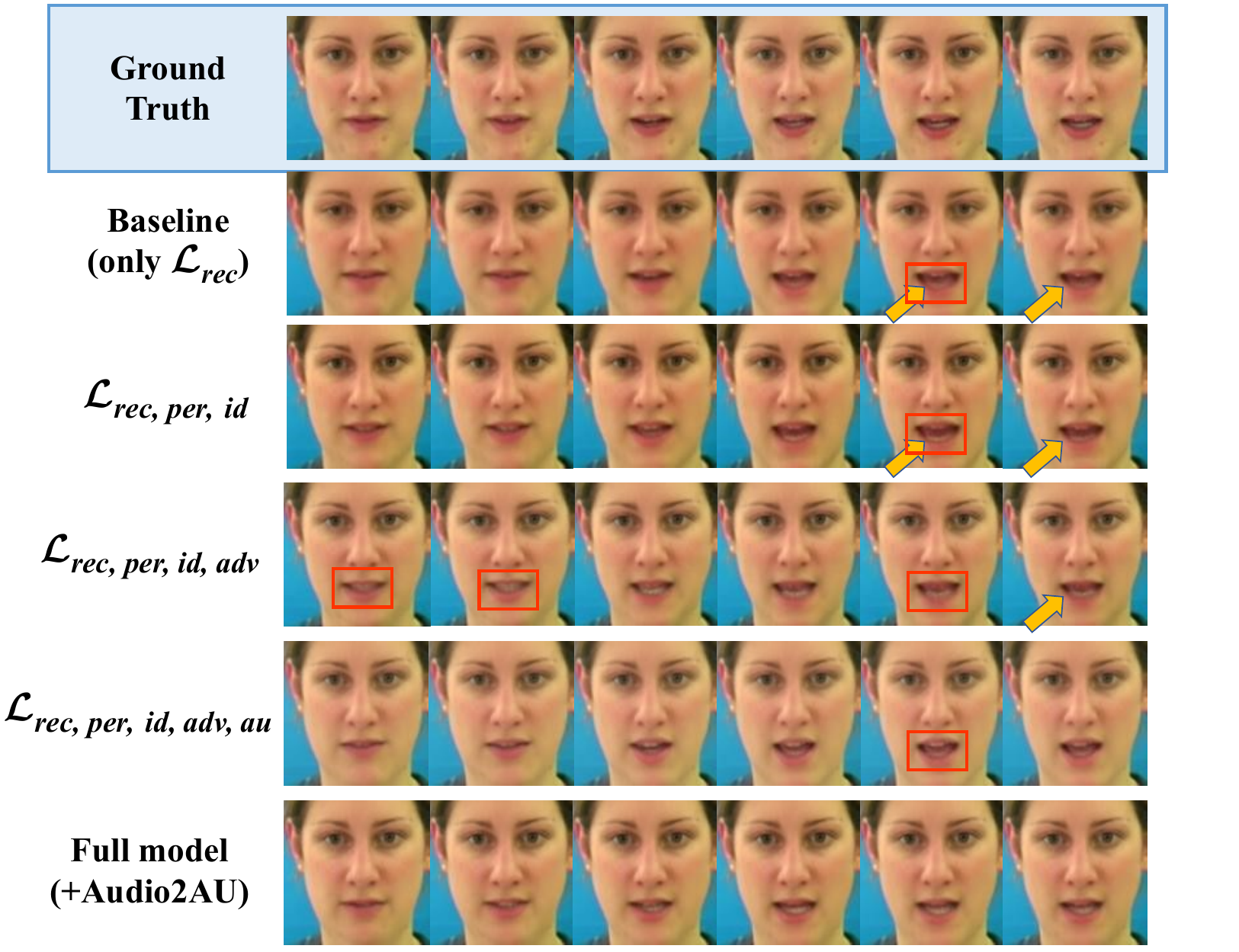}
  \caption{Qualitative results of ablation study on the GRID test set. Each row is a continuous sequence of talking faces. Audio2AU means our proposed Audio-to-AU module.}
  \vspace{-2.5mm}
  \label{fig.5}
\end{figure}

In Fig.~\ref{fig.5}, we also show the visualized examples after adding different components on the GRID test set. Our baseline only uses reconstruction loss, and the generated images are blurry. After adding perceptual loss and identity loss, the image quality is improved, such as facial texture and teeth indicated by the arrows. With the addition of adversarial loss, the facial texture and teeth are further refined. After adding AU loss, lip synchronization has been improved. Finally, the full model with the Audio-to-AU module achieves the best results in both image quality and lip-sync accuracy. We also use red boxes to mark some examples where the lips are not very synchronized in Fig.~\ref{fig.5}.

\subsection{User study}

To subjectively evaluate our model, we further conduct a user study. 15 participants are required to compare the 30 videos generated by Song et al. \cite{song2018talking}, Jamaludin et al. \cite{jamaludin2019you} and our method. The identity images and audios are randomly selected from the GRID and TCD-TIMIT datasets. Users are required to rank videos generated by different methods from the following three perspectives: (1) Lip-sync accuracy; (2) Image quality; and (3) Video fluency, whether the video jitters. We summarize the evaluation results in Table \ref{tab:3}.

\begin{table}[t]
\begin{center}
\small
\setlength\tabcolsep{5pt}
\begin{tabular}{|c|c|c|c|}
\hline
Method&Lip-sync Accuracy&Image Quality& Video Fluency\\
\hline\hline
Ours vs. CRAN \cite{song2018talking}& 0.77 / 0.23 & 0.73 / 0.27 & 0.92 / 0.08  \\
Ours vs. Speech2Vid \cite{jamaludin2019you}& 0.70 / 0.30 & 0.69 / 0.31 & 0.86 / 0.14  \\
\hline
\end{tabular}
\end{center}
\caption{User study on the quality of videos generated by different methods. All the users are required to compare the videos generated by our method and others.}
\vspace{-2.5mm}
\label{tab:3}
\end{table}

As shown in Table \ref{tab:3}, our method is superior to the methods of Song et al. \cite{song2018talking} and Jamaludin et al. \cite{jamaludin2019you} in all aspects, especially in video fluency. The results of the user study further verify the effectiveness of our method in both image quality and lip-sync accuracy.
\vspace{-2.5mm}
\section{Conclusion}
\vspace{-2.5mm}
In this paper, we propose a novel talking head generation system, which uses both audio and speech-related facial action units (AUs) as driving information. The proposed Audio-to-AU module is used to obtain speech-related AU information. AU classifier is used to supervise that the generated frames contain correct AU information. We also utilize frame discriminator for adversarial training to improve the realism of generated frames. We conduct extensive experiments for quantitative and qualitative evaluation, and we use the ablation study to verify each component's contribution to our model. The experimental results on the GRID dataset and TCD-TIMIT dataset demonstrate the superiority over state-of-the-arts in both image quality and lip-sync accuracy. In the future, we will focus on multimodal representation fusion to promote complementation.
\section*{Acknowledgments}

This work is supported by the National Natural Science Foundation of China under Grants of 61503277 and 61771333. 

\bibliography{reference}
\clearpage

\appendix
\section{Supplementary Material}



\noindent We show more interpretations and experiments in the supplementary material. In Section A.1, we introduce the input details of the video stream and audio stream in the model. In Section A.2, we verify the effectiveness of the AU classifier. In Section A.3, we show more visualization results.

\subsection{Data Pre-processing}

\begin{figure}[!htb]
  \centering
  \includegraphics[width=\linewidth]{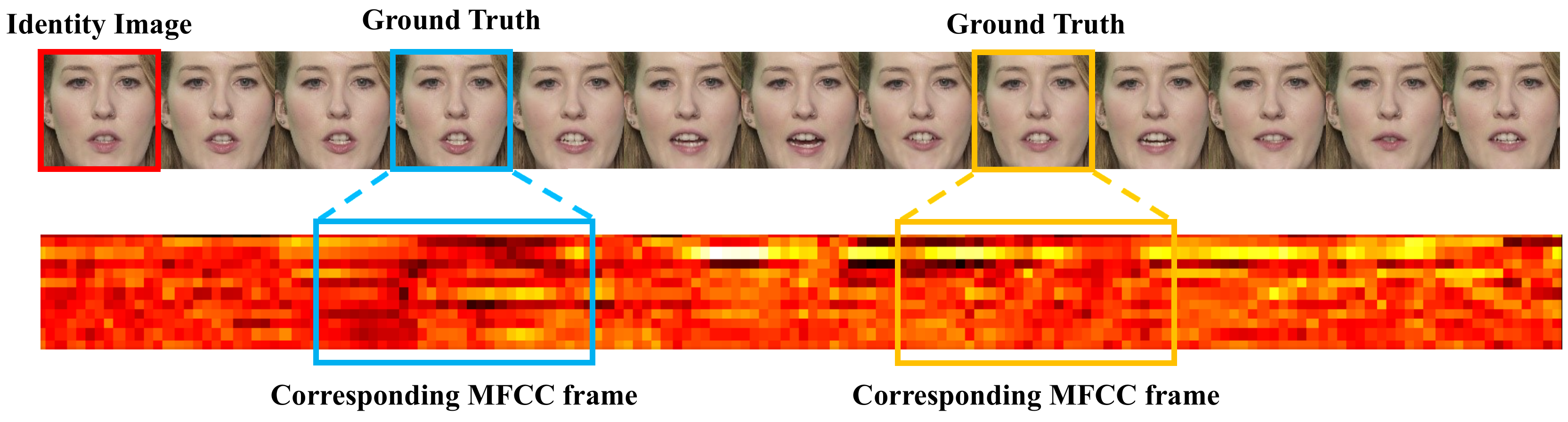}
  \caption{The input details of video stream and audio stream in the model.}
  \label{fig.6}
\end{figure}

As shown in Fig.~\ref{fig.6}, each recording in the GRID \cite{cooke2006audio} dataset and TCD-TIMIT dataset \cite{harte2015tcd} is processed into a video stream and an audio stream. The video stream consists of consecutive cropped video frames. The audio stream consists of consecutive 12-dimensional MFCC features. We use the first frame of each video as the input identity information. For audio input, we slide a fixed window containing 280ms audio information on the audio stream in Fig.~\ref{fig.6}. The corresponding video frame in the middle of the audio sliding window is the ground truth. So we compare the image generated by each audio frame with the corresponding ground truth. 

\begin{table}[!b]
\begin{center}
\begin{tabular}{|c|c|c|c|c|}
\hline
\multirow{2}*{AU Number}&\multicolumn{2}{c|}{Training Set}&\multicolumn{2}{c|}{Test Set}\\
\cline{2-5}
&F1 score& Accuracy& F1 score& Accuracy \\
\hline\hline
AU10 &0.971 &96.81\%&0.875 &88.60\% \\
AU14 &0.896&95.54\%&0.752 &91.07\%\\
AU20  &0.955&95.13\%&0.896 &89.85\%\\
AU25  &0.983&97.16\%&0.979 &96.38\%\\
AU26  &0.920&91.99\%&0.854 &82.18\%\\
\hline
Average  &0.945&95.33\%&0.871 &89.62\%\\
\hline
\end{tabular}
\end{center}
\caption{Results of our pre-trained AU classifier on real images of the GRID training set and test set.}
\label{tab:4}
\end{table}

\subsection{AU Detection with the Pre-trained AU Classifier}

In order to enable the AU classifier to judge whether the generated image contains the correct AU information, we use the GRID dataset to pre-train it. The training data are the images and corresponding speech-related AU labels. We use Openface \cite{baltrusaitis2018openface} to extract the speech-related AU labels of each image. 

The results of our pre-trained AU classifier are shown in Table \ref{tab:4}. The average F1 score of AUs is 0.945 and the average accuracy of AUs is 95.33\% on the training set. This proves that our AU classifier can effectively capture the AU information. On the test set, it achieves 0.871 on average F1 score and 89.62\% on average accuracy, which shows that our pre-trained AU classifier has strong generalization ability.

To further verify the effectiveness of the AU classifier, we use it to detect the speech-related AUs of the images generated by our proposed method. The detection results are shown in Table~\ref{tab:5}. Because the ground truth AU labels are extracted by Openface \cite{baltrusaitis2018openface}, the Openface detection results are accurate. Since the AU classifier is pre-trained on the GRID dataset, when experimenting on the TCD-TIMIT dataset, we need to use the new dataset to refine it to adapt to the new domain. The detection results of the AU classifier on the GRID test set are very close to that of Openface. On the TCD-TIMIT data set, the AU classifier detection results on average F1 score is lower than Openface detection, but accuracy is higher. The main reason is that the distribution of AUs on the TCD-TIMIT dataset is unbalanced, so the prediction of unbalanced AU tends to be absent. Therefore, the wrong results may be predicted as correct by the AU classifier, resulting in lower F1 score and higher accuracy. In general, the AU classifier can judge whether the generated image contains the correct AU information.

\begin{table}[t]
\begin{center}
\begin{tabular}{|c|c|c|c|c|}
\hline
\multirow{2}*{method}&\multicolumn{2}{c|}{GRID}&\multicolumn{2}{c|}{TCD-TIMIT}\\
\cline{2-5}
&Avg. F1 score&Avg. Accuracy &Avg. F1 score& Avg. Accuracy \\
\hline\hline
AU classifier  &0.746 &80.77\%&0.507 &88.56\%\\
\hline
Openface\cite{baltrusaitis2018openface} &0.751 &80.92\%&0.590 &84.92\% \\
\hline
\end{tabular}
\end{center}
\caption{The AU detection results of the images generated by our proposed method on AU classifier and Openface respectively. Because the ground truth AU labels of the real images are extracted by Openface, the Openface detection results are accurate.}
\label{tab:5}
\end{table}

\begin{figure}[t]
  \centering
  \includegraphics[width=\linewidth]{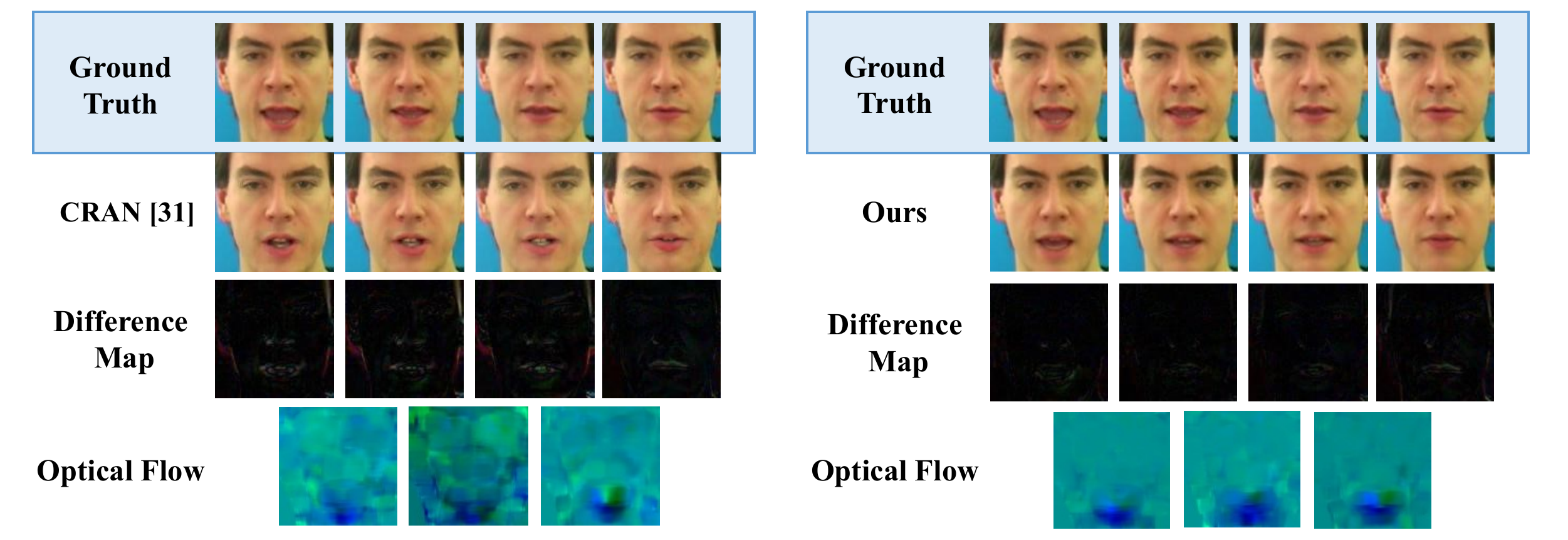}
  \caption{Difference map and optical flow of the images generated by CRAN \cite{song2018talking} and our proposed method.}
  \label{fig.7}
\end{figure}

\begin{figure}[!b]
  \centering
  \includegraphics[width=0.95\linewidth]{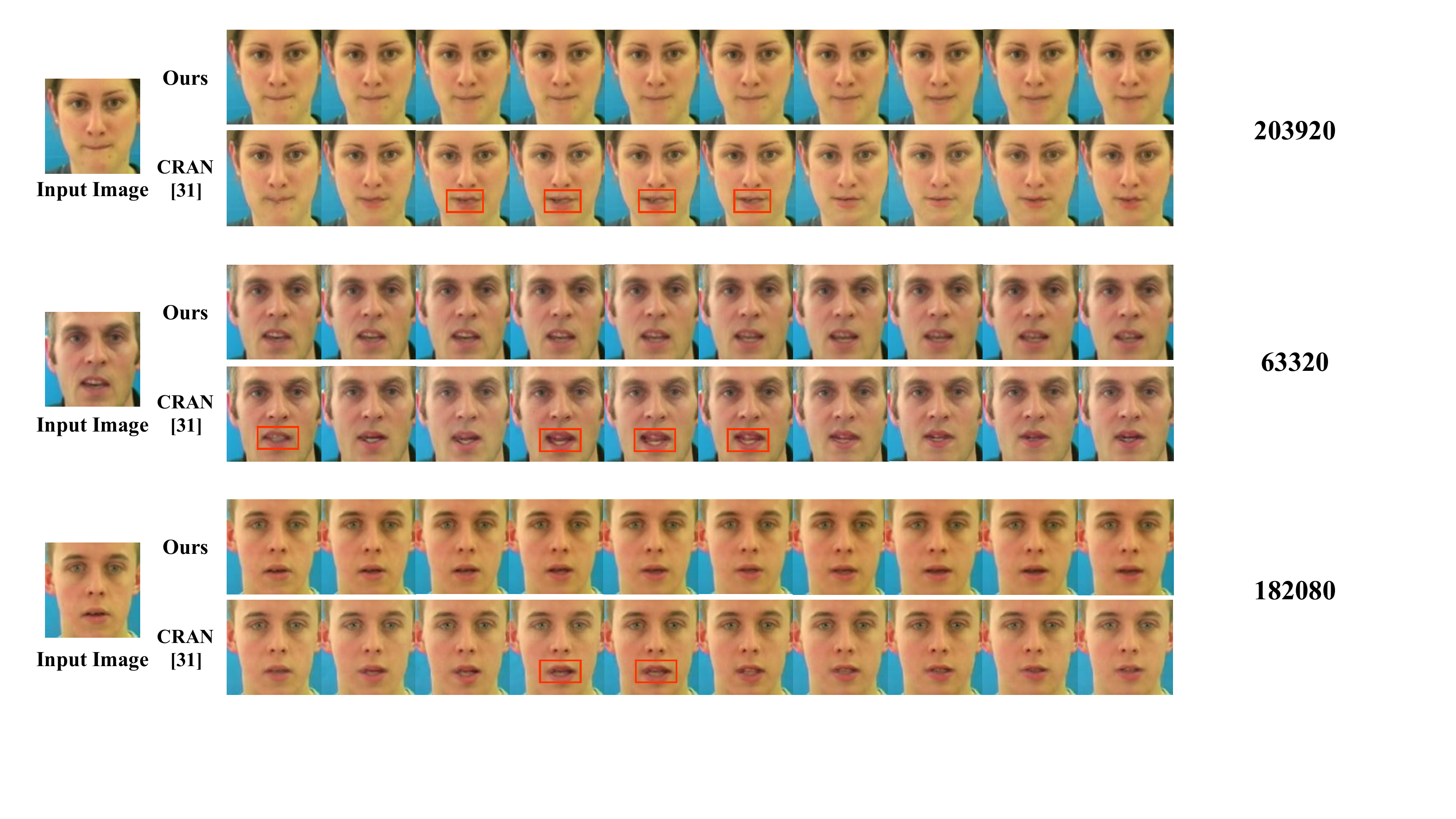}
  \caption{Example of generated videos driven by silent audio with no one speaking and only slight noise. The images generated by our method is consistent with the input image, which obviously suppresses the lip movement.}
  \label{fig.8}
\end{figure}

\begin{figure}[!b]
  \centering
  \includegraphics[width=0.83\linewidth]{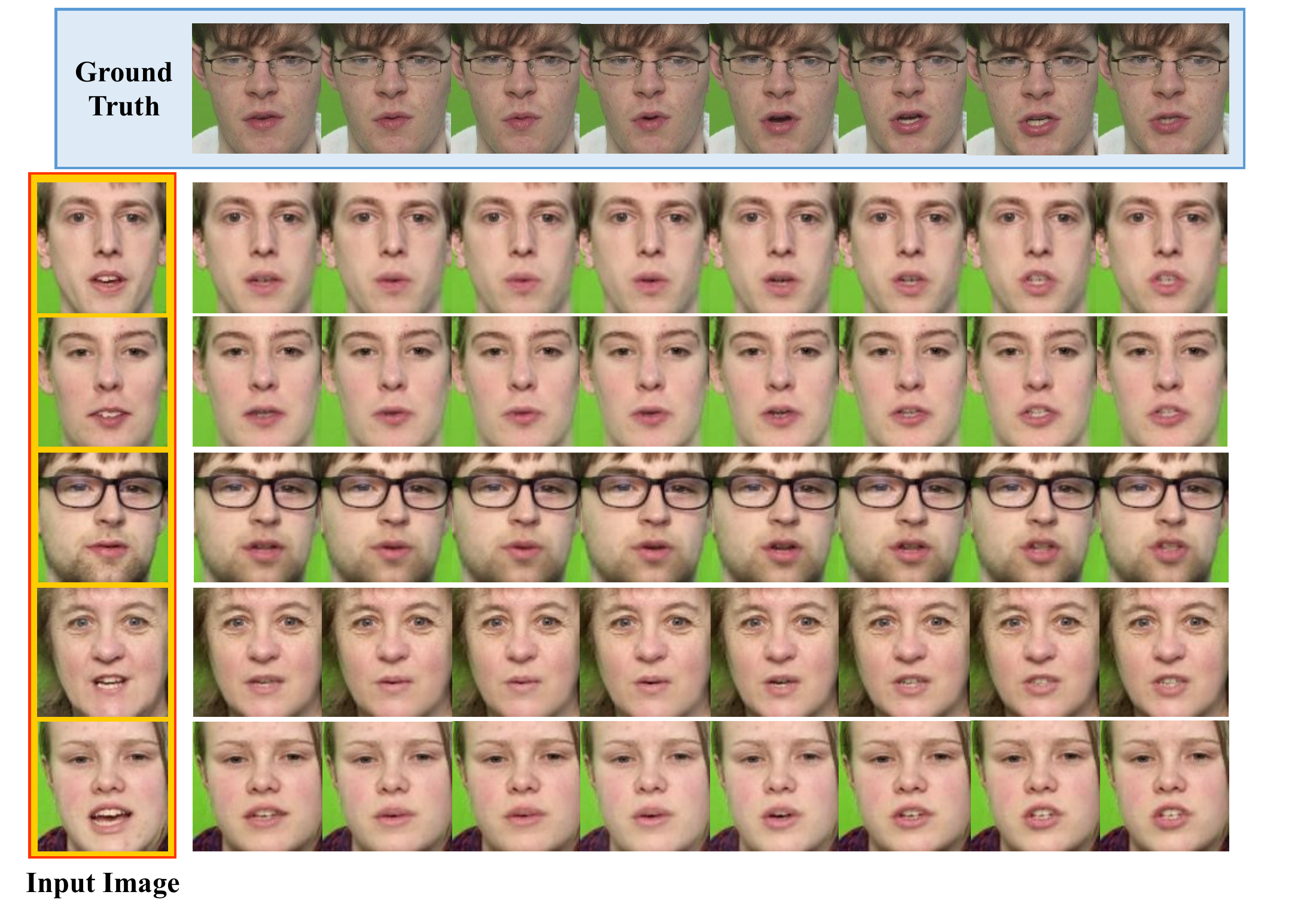}
  \caption{More example results produced using our proposed model on the TCD-TIMIT test set. We use the same audio clip that corresponds to the word "wash" for different speakers to generate talking head videos.}
  \label{fig.9}
\end{figure}

\vspace{-2mm}
\subsection{More Visualization Results}
\vspace{-2mm}
To compare the difference between the generated image and the ground truth image more clearly, we show the difference map in Fig.~\ref{fig.7}. There are obvious differences in the mouth and the edge of the face in the difference map of Song et al. \cite{song2018talking}. The difference map of ours only has a small difference in the mouth region. We also use optical flow to represents the motion between the generated frames. The video generated by Song et al. \cite{song2018talking} is accompanied by jitter, so the optical flow is chaotic. The changes of our video frames are concentrated in the mouth, so the optical flow is very clear.

\begin{figure}[!t]
  \centering
  \includegraphics[width=\linewidth]{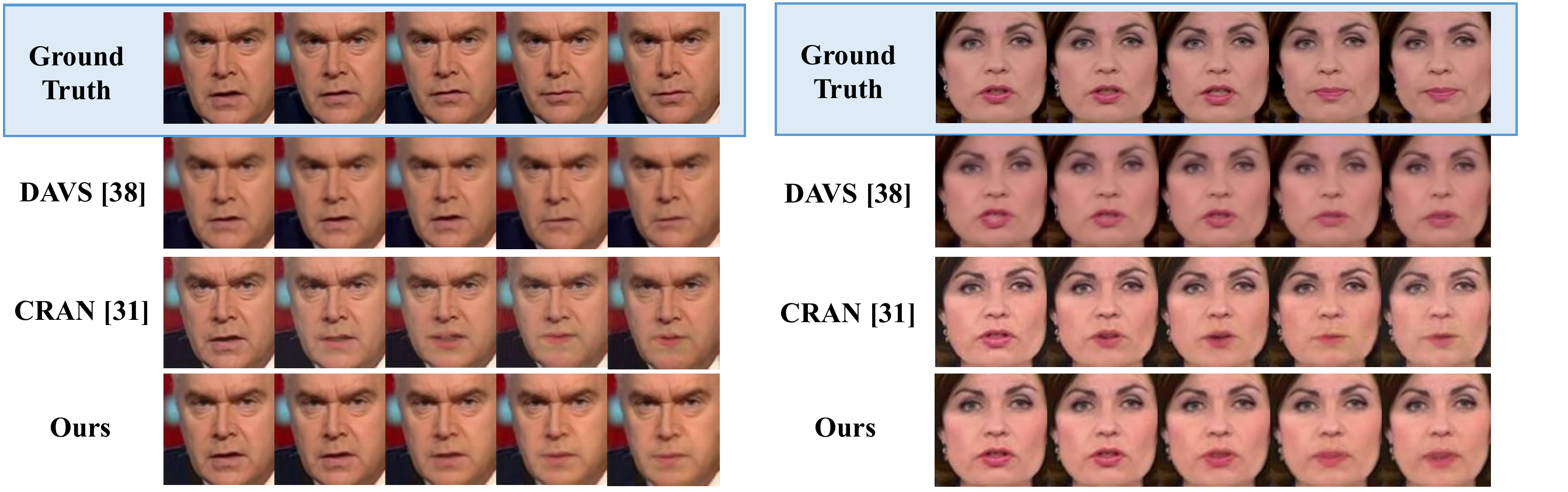}
  \caption{Example of generated frames produced by our proposed model and other methods on the LRW dataset. DAVS \cite{DBLP:conf/aaai/Zhou000W19} is trained on the LRW dataset. CRAN \cite{song2018talking} and our model are trained on the GRID dataset.}
  \label{fig.10}
\end{figure}

When no one is speaking, the mouth should not perform any movement and be consistent with the input. Therefore, we test the effect of our model without anyone speaking. The audio clip only has slight noise. Fig.~\ref{fig.8} shows the generation results of three different subjects using silent audio. The speakers are taken from the GRID test set. The mouth of the first subject is closed in the input image, and the last two subjects are open. We can clearly see that whether the mouth of the input is closed or open, the video generated by our method is consistent with the input image. However, the video generated by Song et al. \cite{song2018talking} can not suppress mouth movement, as shown in the red box.

We also show the results of generated videos for different speakers using the same audio in Fig.~\ref{fig.9}. The speakers are taken from the TCD-TIMIT test set. The content of the audio clip is the word "wash". We can see that the mouth movements generated by our model match the word very well.

To test the generation effect of our proposed model when cross-database, we conduct experiments on the LRW dataset \cite{chung2016lip}. The example of generated frames can be seen in Fig.~\ref{fig.10}. Our proposed model and CRAN \cite{song2018talking} are trained on the GRID dataset, and DAVS \cite{DBLP:conf/aaai/Zhou000W19} are trained on the LRW dataset. It can be seen that the images generated by DAVS are blurry, and the face texture of the images generated by CRAN are not well preserved. Our method can still generate high-quality talking face images when cross-database.

\begin{figure}[!t]
  \centering
  \includegraphics[width=\linewidth]{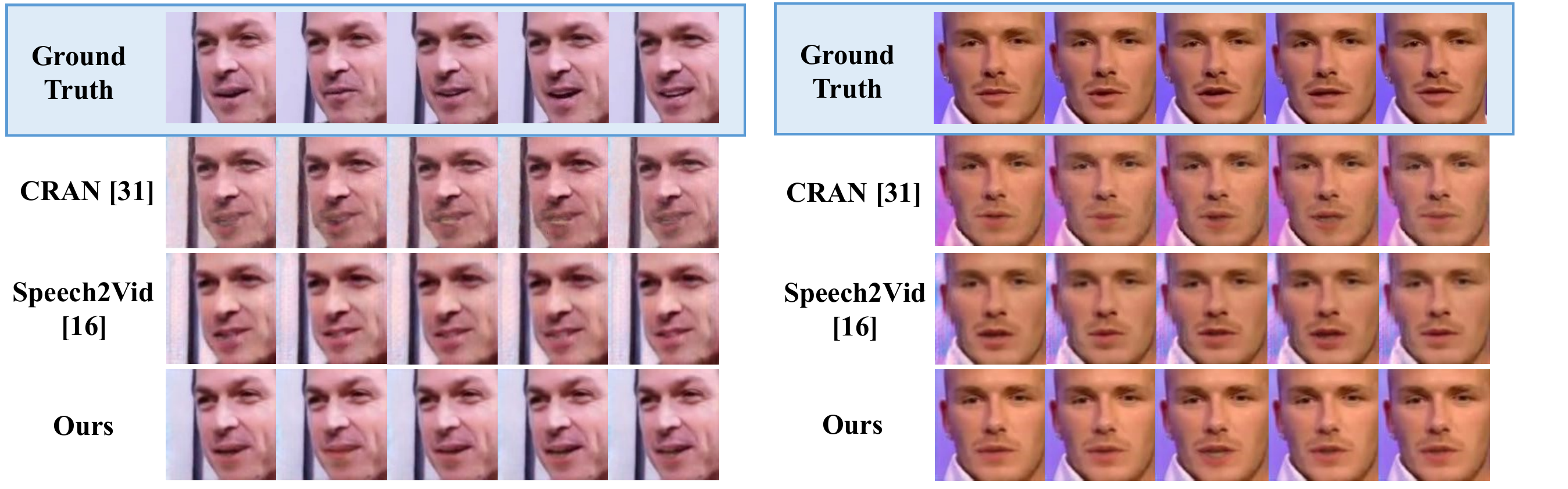}
  \caption{Example of generated frames produced by our proposed model and other methods on the VoxCeleb2 dataset. All of the methods are trained on the GRID dataset.}
  \label{fig.11}
\end{figure}

We also try to use our model trained on the GRID dataset to experiment on the VoxCeleb2 dataset \cite{chung2018voxceleb2}. The example of generated results can be seen in Fig.~\ref{fig.11}. All of the methods are pre-trained on the GRID dataset. Since most of the videos in the VoxCeleb2 dataset have large head movements, they are not suitable for the model trained on the GRID dataset. Therefore, we only choose the examples of the faces with small head movement in the VoxCeleb2 test set. From the Fig.~\ref{fig.11} we can clearly see that the mouths generated by song et al. \cite{song2018talking} are very blurry, and the image quality generated by Jamaludin et al. \cite{jamaludin2019you} is very low, especially the background. The images generated by our method are superior to others in both image quality and lip-sync accuracy.

\end{document}